\documentclass[conference]{IEEEtran}
\usepackage{times}
\usepackage{graphicx}

\usepackage[usenames,dvipsnames,table,xcdraw]{xcolor}
\usepackage{titletoc}
\usepackage{float}
\usepackage{listings}
\usepackage{multirow}
\usepackage{xparse}
\usepackage{optidef}
\usepackage{algorithm}
\usepackage{algpseudocode}
\usepackage{xcolor}
\usepackage{blindtext}
\usepackage{color,soul}
\usepackage{colortbl} 
\usepackage{hyperref}

\usepackage[numbers, sort&compress]{natbib}
\usepackage{multicol}

\usepackage{hyperref}
\hypersetup{
    colorlinks=true,
    linkcolor=MidnightBlue,
    filecolor=MidnightBlue,      
    urlcolor=MidnightBlue,
    citecolor=MidnightBlue,
}

\usepackage{url}
\usepackage{booktabs} 
\usepackage{amsmath}
\usepackage{float} 
\usepackage{stfloats}
\usepackage{afterpage}
\usepackage{svg}
\usepackage{authblk}

\usepackage{titlesec}

\usepackage[font=footnotesize,labelfont=bf,tableposition=top]{caption}
\usepackage{subcaption}
\usepackage{graphicx} 

\begin{document}

\newcommand{\klin}[1]{{\color{blue}{Kevin: #1}}}

\title{
  {\fontsize{24}{29}\selectfont Consistency Policy}\\[0.2em] 
  {\fontsize{19}{21}\selectfont Accelerated Visuomotor Policies via Consistency Distillation} 
}

\makeatletter
\renewcommand\AB@affilsepx{, \protect\Affilfont}  
\makeatother


\author{
Aaditya Prasad$^1$, Kevin Lin$^1$, Jimmy Wu$^2$, Linqi Zhou$^1$, Jeannette Bohg$^1$ \\ 
$^1$Stanford University \quad
$^2$Princeton University
\\ \href{https://consistency-policy.github.io}{https://consistency-policy.github.io}
}


%

\maketitle
\begin{abstract}
Many robotic systems, such as mobile manipulators or quadrotors, cannot be equipped with high-end GPUs due to space, weight, and power constraints. These constraints prevent these systems from leveraging recent developments in visuomotor policy architectures that require high-end GPUs to achieve fast policy inference. In this paper, we propose Consistency Policy, a faster and similarly powerful alternative to Diffusion Policy for learning visuomotor robot control. By virtue of its fast inference speed, Consistency Policy can enable low latency decision making in resource-constrained robotic setups. A Consistency Policy is distilled from a pretrained Diffusion Policy by enforcing self-consistency along the Diffusion Policy's learned trajectories. We compare Consistency Policy with Diffusion Policy and other related speed-up methods across 6 simulation tasks as well as three real-world tasks where we demonstrate inference on a laptop GPU. For all these tasks, Consistency Policy speeds up inference by an order of magnitude compared to the fastest alternative method and maintains competitive success rates. We also show that the Conistency Policy training procedure is robust to the pretrained Diffusion Policy's quality, a useful result that helps practioners avoid extensive testing of the pretrained model. Key design decisions that enabled this performance are the choice of consistency objective, reduced initial sample variance, and the choice of preset chaining steps.
\end{abstract}

    


\IEEEpeerreviewmaketitle

\section{Introduction}
Diffusion models have recently demonstrated impressive results in Imitation Learning for robot control \cite{Chi-RSS-23, reuss2023goal, wang2023diffusion, octo_2023, ze20243d}. In particular, Diffusion Policy \cite{Chi-RSS-23} demonstrates state-of-the-art imitation learning performance on a variety of robotics tasks.

One key drawback of diffusion models is the inference time required to generate actions. Diffusion models produce outputs by sequentially denoising from an initial, noisy state. This process means that they require multiple forward evaluations to predict an action and that reducing the number of evaluations degrades performance. Diffusion Policy \cite{Chi-RSS-23} uses a diffusion framework named {\em Denoising Diffusion Probabilistic Models\/} (DDPM) \cite{ho2020denoising} evaluated using 100 denoising steps, which on an NVIDIA T4 can take around one second per action generation. 

Such slow inference constrains the use cases for Diffusion Policy to tasks and settings that tolerate lengthy reaction times and high computational costs. While quasi-static tasks such as simple pick-and-place or part assembly can permit slow inference speeds, dynamic tasks such as balancing objects or navigating dynamic environments often require faster control frequencies. Furthermore, Diffusion Policy can be impractically slow for robots with on-board compute constraints. Given these observations, \textbf{our goal is to retain the performance of Diffusion Policy while drastically reducing inference time.}

In the image generation domain, there has been much interest in distillation techniques \cite{yin2023one,liu2023instaflow,salimans2022progressive,song2023consistency,song2023improved} that use a pre-trained diffusion model to teach a new student model how to take larger denoising steps, reducing the total number of function evaluations required for generation. One set of distillation techniques \cite{kim2023consistency, song2023consistency, song2023improved} is based on the insight that a trained diffusion model can be interpreted as solving an ODE \cite{song2021scoresde}.
These approaches use the uniqueness of the solutions to these ODEs and enforce consistency between denoising steps that begin at different positions on the same ODE trajectory. The distilled student network is thus called a Consistency Model. In image generation, distilled Consistency Models have been shown to produce single or few-step generations that rival traditional diffusion models in sample quality.
\begin{figure}
    \centering

    \begin{subfigure}{0.97\columnwidth}
        \centering
        \includegraphics[width=\textwidth]{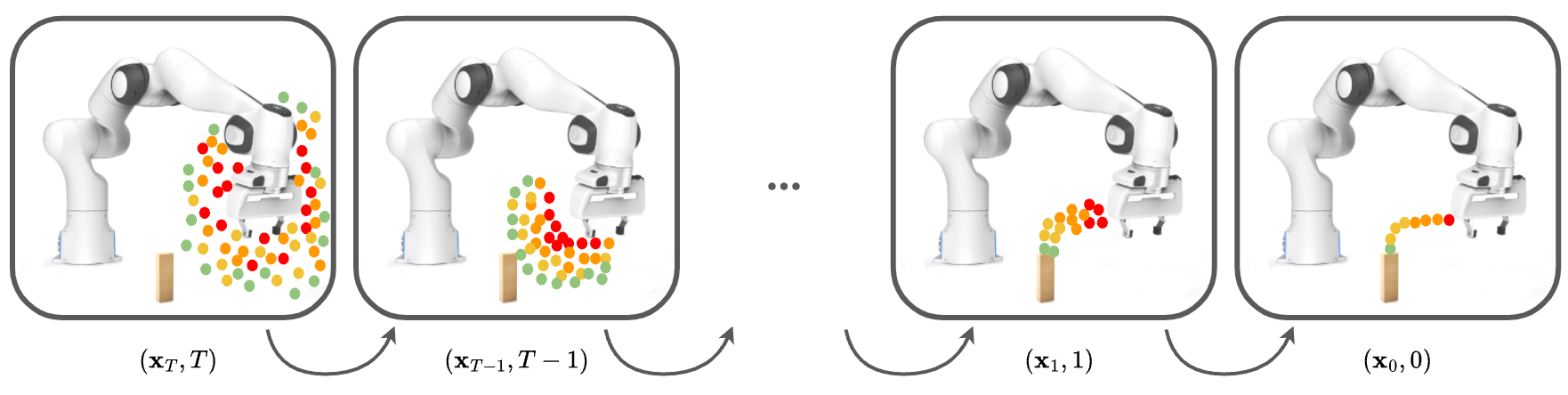}
        \caption{\textbf{Diffusion Policy}}
        \label{fig:dp}
    \end{subfigure}

    \vspace{5mm} 

    \begin{subfigure}{0.97\columnwidth}
        \centering
        \includegraphics[width=\textwidth]{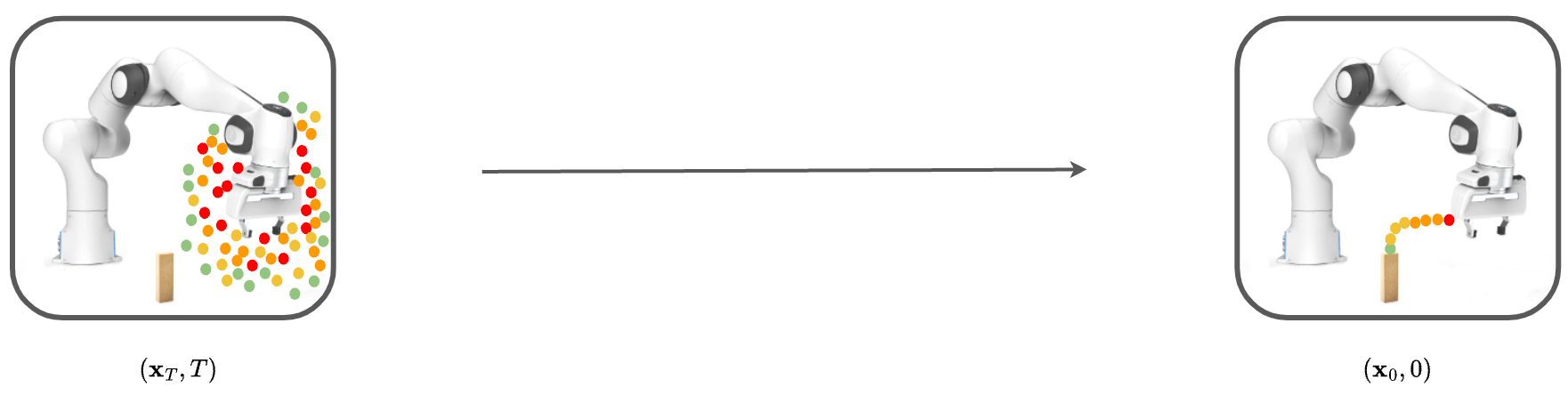}
        \caption{\textbf{Consistency Policy}}
        \label{fig:cp}
    \end{subfigure}



    \caption{Both Diffusion and Consistency Policy work by sampling random actions and denoising them into predictions of actions. $x_t$ denotes the current action distribution at a time $t \in [0, T]$, where larger times correspond to noisier actions. The figure shows distributions of predicted action sequences (indicated by the sequences of red to green dots) at different stages of the respective generation process. a) \textbf{Diffusion Policy} denoises an action sequence over many steps, resulting in high inference costs when deploying the policy on a robot. b) \textbf{Consistency Policy} generates an action sequence in a single step, allowing for much faster inference speeds than Diffusion Policy while retaining competitive success rates.}
    \label{fig:both-images}
\end{figure}
We adapt these Consistency Model frameworks for the robotics domain. 
We first replace the diffusion frameworks that Diffusion Policy employs with EDM \cite{Karras2022edm}, an analogous multi-step framework more commonly used for consistency distillation.
We train a teacher model using the EDM framework and then distill it using an adaptation of the \textit{Consistency Trajectory Model} (CTM) objective proposed by \citet{kim2023consistency}. Key design decisions include the specific choice of consistency objective, reduced initial sample variance, and the choice of preset chaining steps. We also provide insights about the role of dropout along a specific region of the CTM objective, and analysis of Consistency Policy's robustness to teacher quality. 

Overall, we demonstrate that inference speed of our approach is on average about an order of magnitude faster than the fastest baseline (see Table \ref{tab:average_reward_combined}) and maintains similar or higher success rates than all baselines on a variety of tasks. 

\section{Related Work}
Diffusion models have achieved many state-of-the-art results across image, audio, video, and 3D generation \cite{song2021denoising, kim2023consistency, poole2023dreamfusion, bar2024lumiere, sheynin2023emu, ruan2023mm}. In the context of robotics, diffusion models have been used as policy networks for imitation learning to great effect \cite{Chi-RSS-23, wang2023diffusion, octo_2023}. 
However, vanilla variants of diffusion models, such as {\em Denoising Diffusion Probabilistic Models\/} (DDPMs) \cite{ho2020denoising}, suffer from long inference times due to their need for many iterative sampling steps. 
In particular, DDPM \cite{ho2020denoising} can be interpreted as solving a Stochastic Differential Equation backwards in time, and is thus characterized by a stochastic denoising process that integrates small amounts of Brownian motion as it generates an output. DDPM has a fixed step count that is often 100+, making it the slowest framework used by Diffusion Policy~\cite{Chi-RSS-23}. 

One line of work \cite{song2021denoising} addresses the long inference times of diffusion models by reducing the number of denoising steps required for a prediction.
As opposed to the stochastic solver of DDPM, \emph{Denoising Diffusion Implicit Models} (DDiM) \cite{song2021denoising} can be interpreted as integrating over a deterministic ODE \cite{salimans2022progressive}. 
Importantly, DDiM allows for a variable step count, meaning a DDiM-based network could be trained with a large number of denoising steps but evaluated at inference time with a much smaller number of steps. EDM \cite{Karras2022edm} follows this pattern of integrating the deterministic ODE and also allows for a small number of denoising steps at test time. EDM differs from DDiM through modifications to preconditioning and weighting.
However, even with variable step schemes such as DDiM and EDM, reducing the number of denoising steps at inference time often reduces sample quality.

Another line of work, introduced by \citet{shih2023parallel}, aims to speed up diffusion models through parallel sampling. Instead of denoising sequentially over the ODE, ParaDiGMS \cite{shih2023parallel} attempts to converge sliding batches of points along a diffusion ODE's trajectory in parallel via Picard Iteration on those points.
This method has the potential for large speed-ups as points can converge long before a sequential solver may have reached them, allowing for all previous points to be skipped. 
However, ParaDiGMS drastically increases memory requirements due to this parallelization. In robotics settings, processing hardware is often limited and available compute is constrained by other processes that need to run in parallel to the policy network. An actual user might see speed gains from ParaDiGMS drop rapidly as they are forced to lessen their batch window size by VRAM availability or other constraints. Additionally, this method still remains slower than single step prediction, which Consistency Policy enables. 

Distillation based techniques \cite{salimans2022progressive, meng2023distillation} have also been explored to accelerate diffusion model inference speeds in the text-to-image domain. 
Many of these distillation techniques start with a pretrained teacher model and train a new student model to take larger steps over the ODE trajectories that the teacher has already learned to map \cite{yin2023one, liu2023instaflow, salimans2022progressive, kim2023consistency, song2023consistency, song2023improved}. By taking these larger steps, the student model is able to complete a generation in a smaller total number of steps. 

Of the distillation based techniques, the consistency model works \cite{song2023consistency, song2023improved, kim2023consistency} support both single and multi-step sampling of outputs. 
Consistency distillation techniques exploit the \textit{self-consistency property} \cite{song2023consistency} of ODE trajectories by training the student model to predict the same output when given two distinct points along the same ODE trajectory. This objective was first introduced by \citet{song2023consistency}, who choose a pair of adjacent input points and taught the student model to map those input points to the same starting point on the given ODE trajectory, where the distinct points are sampled with the help of a pretrained diffusion model. \citet{kim2023consistency} generalized this method by training for arbitrary step sizes and arbitrarily spaced input points, achieving state of the art results in the image-generation domain. In this work, we study how the latter, more generalized framework, {\em Consistency Trajectory Models\/} (CTMs) \cite{kim2023consistency}, can be adapted for the robotics domain.

Concurrently, \citep{anonymous2024consistency, chen2023boosting} have explored the use of consistency models as a policy class for state-based continuous control. \citet{chen2023boosting} adapts the consistency distillation model objective \cite{song2023consistency} for state-based offline RL settings. The authors do not use the generalized CTM framework from \citet{kim2023consistency} and are also unable to directly distill a teacher model because of their focus on Q-learning rather than behavior cloning. \cite{anonymous2024consistency} leverages the consistency training (as opposed to distillation) objective from \cite{song2023consistency} for state-based, continuous control tasks. The consistency training objective replaces a teacher model with a Monte-Carlo estimator and thus allows for teacher-free training.
While it may achieve good results for state-based policies on common RL benchmarks, we demonstrate that this consistency training objective does not lead to an adequate success rate for much more high-dimensional {\em image\/}-based policies on relatively more complex robotics tasks. 

Finally, there has been a long line of work using non-diffusion based model architectures for visuomotor robotics policies. Such alternatives often perform worse than diffusion policies on the same tasks, or require external computational resources that may be unavailable in many robotics settings.

The original Diffusion Policy paper \cite{chi2023diffusion} benchmarked against prior state of the art imitation learning (IL) algorithms on several robotics simulation \cite{mandlekar2021matters, florence2021implicit, shafiullah2022behavior} and real world tasks. Because Diffusion Policy outperformed than all of these baselines, we choose to baseline against only Diffusion Policy and other Diffusion Policy inference acceleration methods. Most notable among these inferior baselines was Behavioral Transformer \cite{shafiullah2022behavior}, which represents a key alternative to Diffusion Policies: single-step transformer-based models \cite{brohan2022rt, brohan2023rt}

RT-1 \cite{brohan2022rt} is a strong single-step transformer-based baseline but was designed for massively scaled pre-training. Rather than Diffusion Policy \cite{chi2023diffusion} and our own Consistency Policy, diffusion policy networks such as Octo \cite{octo_2023} are more readily comparable to RT-1. In fact, Octo includes RT-1 as a baseline and shows a marked improvement over it as well as another transformer baseline. Since the improvements we introduce in this paper are orthogonal to the specific diffusion policy formulation, an Octo policy could also be distilled and made into a single or few-step policy network. This could be an interesting direction for future work.

Furthermore, we decided not to baseline against RT-2 \cite{brohan2023rt} and other Vision or Large Language Model enabled policy networks \cite{zengalarge, li2023vision} because they leverage the vast pretraining and scale of the integrated language model and have to be run in the cloud rather than with an on-board computer (which is the setting we are considering in this work).

\section{Consistency Policy}


We formulate a visuomotor robot policy as a \textit{Consistency Trajectory Model} \cite{kim2023consistency}, and denote this as a Consistency Policy. 
In this section, we begin with a short introduction to Diffusion Models. We then describe how to train a Consistency Policy, which requires training a teacher Diffusion Policy and then distilling this teacher model into a Consistency Policy. We then explain our inference procedures, which include a single-step process for the fastest inference time possible as well as a 3-step process that trades off some inference speed for greater accuracy. Finally, we cover some implementation details. 

\subsection{Preliminaries}

This section provides a gentle introduction to diffusion models as we used them, so we focus on the ODE interpretation of these models. For further reading, see \cite{ho2020denoising, song2021scoresde, Karras2022edm}. Throughout this paper, the word "trajectory" will only refer to the ODE trajectory parameterized by the diffusion step, which we explain below. Robot motions (either predicted or demonstrated by an expert) will be referred to as ``actions" or ``action sequences", not trajectories.

Our diffusion models learn to map random actions $\mathbf{x}_T$ sampled from the unit Gaussian $\mathcal{N}(0, \mathbf{I})$ to specific actions $\mathbf{x}_0$ drawn from the expert action distribution conditioned on the current observation (which we denote as $p_0(\mathbf{x} | o)$). The subscript $t$ with $0 \leq t \leq T$ refers to time along the trajectory that maps a point from the simple Gaussian distribution at time $T$ to the complex data distribution at time $0$.


This process is often formulated as a {\em Probability Flow ODE\/} (PFODE) \cite{song2021scoresde}, where evolving the PFODE forward in time noises the action and evolving backwards in time denoises the action. A fully denoised action is the policy's prediction of the expert action. 

The general form of this PFODE is 
\begin{equation}
\mathrm{d} \mathbf{x}_t=\left[\boldsymbol{\mu}\left(\mathbf{x}_t, t\right)-\frac{1}{2} \sigma(t)^2 \nabla \log p_t\left(\mathbf{x}_t | o\right)\right] \mathrm{d} t 
\end{equation}
where 
$\boldsymbol{\mu}(\cdot, \cdot)$ is the drift coefficient, $ \sigma(\cdot)$ is the diffusion coefficient, and $p_t(\mathbf{x}_t|o)$ is the noised probability distribution at some time $t \in [0, T]$.  To make $p_T(\mathbf{x}_t | o)$ approach a normal distribution for $T$ sufficiently large, \citet{Karras2022edm} set $\boldsymbol{\mu}(\mathbf{x_t}, t) = 0$ and $\sigma(t) = \sqrt{2t}$.

The gradient of the noised probability distribution, $\nabla \log p_t\left(\mathbf{x}_t|o\right)$, is known as the score. This score function is often intractable to compute, so we approximate it with a neural network. Thus, a denoising step requires evaluating the score function approximator at the current position $\mathbf{x}_t$ and then integrating the resulting $\mathrm{d}\mathbf{x}_t$ using a numerical integration technique. 

Training the score function approximator can be done with numerous objectives, but they all require first performing the forward diffusion process (noising) on samples from the original training data set. The unnormalized, perturbed distribution at timestep t, $p_t(\mathbf{x}_t | o)$, is equivalent to the original data distribution $p_0(\mathbf{x}| o)$ convolved with $\mathcal{N}(0, t^2\mathbf{I})$. Sampling a specific $\mathbf{x}_t \sim p_t(\mathbf{x}_t | o)$ can be done by sampling random noise $\epsilon \sim \mathcal{N}(0, \mathbf{I})$, multiplying it by t, and then adding it to a sample from the original distribution $\mathbf{x} \sim p_0(\mathbf{x}| o)$: $\mathbf{x}_t = \mathbf{x} + t * \epsilon$. Before we pass this position into the score function approximator, we normalize it so it has unit variance.

\subsection{Training}
To train a model capable of few or single-step generation, we begin by training a teacher model and then distill it into a student model. The teacher and student formulations are described below. 
\subsubsection{Teacher Model (EDM)}
The teacher model, which we denote by $s_\phi$, is trained as per the EDM framework \cite{Karras2022edm}. A trained EDM model takes as input the current position $\mathbf{x_t}$ and time $t$ along a PFODE, as well as the conditioning $o$, and is used to estimate the derivative of the PFODE's trajectory:
\begin{equation}
\label{integrating edm}
\frac{\mathrm{d} \mathbf{x}_t}{\mathrm{~d} t}=-\frac{(\mathbf{x}_t - s_\phi\left(\mathbf{x}_t, t;o\right))}{t} 
\end{equation}
An EDM model has to be used alongside a numerical integration method to actually compute positions $\mathbf{x}$ along the PFODE's trajectory. This repeated estimation of the derivative of the ODE followed by its numerical integration is what causes the slow inference speed of Diffusion Models.

Following~\cite{Karras2022edm}, we optimize the {\em Denoising Score Matching\/} (DSM) loss to train the EDM model: 

\begin{equation}
\label{DSM loss}
\mathcal{L}_{\mathrm{DSM}}(\boldsymbol{\theta})=\mathbf{E}_{t, \mathbf{x}_0, \mathbf{x}_t \mid \mathbf{x}_0}[d(\mathbf{x}_0, s_{\boldsymbol{\phi}}\left(\mathbf{x}_t, t ; o\right))]
\end{equation}
The DSM objective takes a sampled point along a PFODE, $(\mathbf{x}_t, t)$, and teaches the EDM model to predict the ground truth initial position $\mathbf{x_0}$. The metric $d(\cdot, \cdot)$ we use is the pseudo-huber loss:
\begin{equation}
d(\boldsymbol{x}, \boldsymbol{y})=\sqrt{\|\boldsymbol{x}-\boldsymbol{y}\|_2^2+c^2}-c
\end{equation}
where $c>0$ is a small constant. We follow Song's \cite{song2023improved} recommendation to set $c = 0.00054\sqrt{D}$ for $D$ dimensional data. This metric acts as a bridge between the standard $l_1$ and $l_2$ norms, and handles outliers more effectively than the $l_2$ loss originally used in EDM. 

Following \citet{Karras2022edm}, we use Heun's second order solver for our numerical integration scheme. We also maintain the timestep discretization scheme described in EDM \cite{Karras2022edm} throughout our work.

\subsubsection{Student Model (Consistency Policy)}


\citet{kim2023consistency} propose a training objective to distill a teacher model $s_\phi(\mathbf{x}_t, t; o)$ into a student model $g_\theta(\mathbf{x}_t, t, s; o)$ and achieve state of the art results on image generation tasks with only one or a small number of inference steps. 
The student model $g_\theta(\mathbf{x}_t, t, s; o)$ is a neural network that takes in a position $\mathbf{x}_t$ along a PFODE, the time $t$, and the observation $o$. The student model learns to output an estimate of $\mathbf{x}_s$ where $s$ is any earlier time along the PFODE.
The student model is trained using a combination of two objectives: the DSM loss (see Eq \ref{DSM loss}) and the CTM loss~\cite{kim2023consistency}, which we now explain in more detail.

\begin{figure}
    \centering
    \includegraphics[width=.5\textwidth]{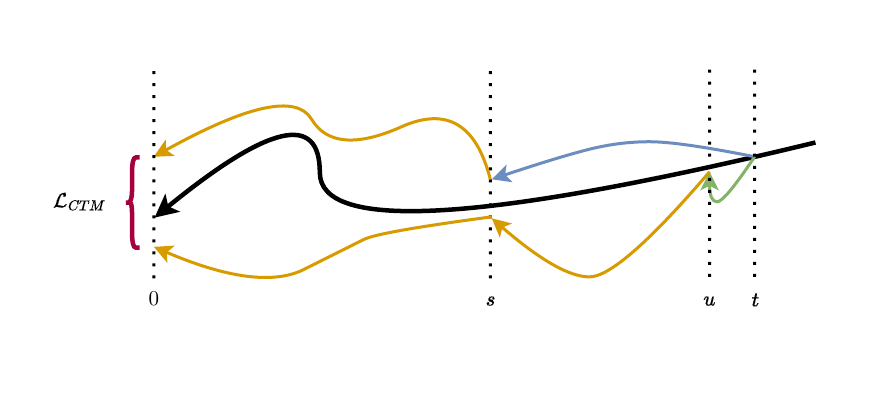}
    \caption{CTM enforces self-consistency along a PFODE (black) by sampling points $s, u, t$ in time such that $0 \leq s < u < t \leq T$, denoising from $t \rightarrow u$ with a teacher model under \textbf{stopgrad} (green), denoising from $t \rightarrow s$ with the student model (blue), and denoising $u \rightarrow s$ with the student model under \textbf{stopgrad}
    (orange). We then use the \textbf{stopgrad} student model to take both generated positions at time $s$ back to time 0. The difference between these two final generations is the computed loss, $\mathcal{L}_{CTM}$ (red). In our experiments, we found $u = t-1$ and $s$ arbitrary below $u$ to work best.}
    \label{fig:training}
\end{figure}

Intuitively, the CTM objective can be understood as enforcing self-consistency along the PFODE since different points $(\mathbf{x_t}, t)$ and $(\mathbf{x_u}, u)$ on the same PFODE should be reconstructed into the same position $\mathbf{x}_s$ at some time $s$ with $0 \leq s < u < t \leq T$. More formally, the CTM objective involves sampling two positions $\mathbf{x}_{t}, \mathbf{x}_{u}$ on the same PFODE and denoising both positions back to the same timestep $s$. After computing $g_\theta(\mathbf{x}_t, t, s; o)$ and $g_\theta(\mathbf{x}_u, u, s; o)$, both of these samples, which we refer to as $\mathbf{x}_s^{(t)}$ and $\mathbf{x}_s^{(u)}$ respectively, are brought back to time 0 using $g_\theta(\mathbf{x}_s^{(t)}, s, 0; o)$ and $g_\theta(\mathbf{x}_s^{(u)}, s, 0; o)$ (see Fig. \ref{fig:training}). This is done before we compute the loss to ensure that the loss metric is always calculated in the fully denoised action space, and is taken from \cite{kim2023consistency}. Thus:
\begin{equation}
\label{CTM Loss}
\mathcal{L}_{CTM} = d(g_\theta(\mathbf{x}_s^{(t)}, s, 0; o), g_\theta(\mathbf{x}_s^{(u)}, s, 0; o))
\end{equation}
where 
\begin{equation}
    \label{t to s}
    \mathbf{x}_s^{(t)} = g_\theta(\mathbf{x}_t, t, s; o)
\end{equation}
\begin{equation}
        \mathbf{x}_s^{(u)} = g_\theta(\mathbf{x}_u, u, s; o)
\end{equation}
and $s, u, t$ are all points on the discretized time mesh. 

The only operation in the prior three equations that we do not put under \textbf{stopgrad}\footnote{\textbf{stopgrad} is a function in automatic differentiation frameworks such as Pytorch which prevents an operation from being added to the computation graph. In Fig. \ref{fig:training}, gradients from the loss are only calculated with respect to the operation from $t \rightarrow s$ (blue). Differentiating with respect to every operation could lead to unstable training and slow or even failed convergence.} is the generation from $t \rightarrow s$, which is described in Eq \ref{t to s}. 

We sample training times $t$ from a uniform distribution over the discretized timesteps. After $\mathbf{x_t}$ is sampled from $\mathcal{N}(0, t^2\mathbf{I})$, $\mathbf{x_u}$ is sampled using $t-u$ steps of the teacher EDM model and the chosen numerical integrator, as per Eq. \ref{integrating edm}. Distillation signal is thus provided via the teacher model's prediction of $\mathbf{x_u}$ given $\mathbf{x_t}$. 







We add this consistency term to the DSM loss in Eq~\ref{DSM loss} to make the final training objective:

\begin{equation}
    \label{total loss}
\mathcal{L}_{CP} = \alpha\mathcal{L}_{CTM} + \beta\mathcal{L}_{DSM}
\end{equation}

with tunable hyperparameters $\alpha$ and $\beta$ and where subscript $CP$ stands for Consistency Policy.

In practice, adapting the standard sampling scheme from \citet{song2023consistency} such that $t$ and $u$ are adjacent timesteps seemed to work the best. We explore this in more detail in Table \ref{tab:consistency-objective-ablation}. Details such as the skip connection and samplers are maintained as in \cite{kim2023consistency}.

\subsection{Inference}
An important property of Consistency Policy is the ability to trade speed for accuracy at inference time without further training of the model. We thus describe two procedures: single-step inference for when speed is paramount and 3-step inference for when more accuracy is desired. Both of these methods remain faster than prior works. 

Single-step inference from our trained Consistency Policy works as follows: sample the initial position $\mathbf{z} \sim \mathcal{N}(0, \mathbf{I})$, compute $\mathbf{x} = g_\theta(z, T, 0; o)$ where T is the max timestep we use during training and $o$ is the current observation, and deploy $\mathbf{x}$ as our action to our environment. Note that we are sampling $\mathbf{z} \sim \mathcal{N}(0, \mathbf{I})$ as opposed to $\mathbf{z} \sim \mathcal{N}(0, 
T^2\mathbf{I})$, which is the standard unnormalizing initial sample. This change pushes our initial point to start much closer to the mean of the normal distribution and empirically performed better than the standard sampling scheme, as we display in Table \ref{tab: initial variance}. An interpretation for this is that sampling closer to the mean ensures that the trajectory is more in-distribution and prevents outliers. This may be related to \citet{pearce2022imitating}'s hypothesis that in imitation learning tasks, it is detrimental to push outputs away from high-likelihood unconditional areas that lie at the center of the expert data distribution, even if such forces are useful in image generation tasks (in the authors' case, with classifier free guidance). 

We perform 3-step generation by chaining generations together as in Consistency Models \cite{song2023consistency}. Given chaining timesteps $\{t_1, t_2\}$, we denoise from $T \rightarrow 0$ as usual, then noise to time $t_1$, denoise back to time 0, and repeat the preceding 2 steps for the remaining chaining timestep. This back-and-forth process can be interpreted as refining the initial prediction. 

These chaining timesteps are hyperparameters. To our knowledge, the original Consistency Models work tuned these steps separately for every task and dataset. Such tuning can become complicated in the robotics setting when real world trials are required to gauge success rates and practitioners may want a strong recommendation that works out of the box.  

Prior works in the image diffusion domain \cite{choi2022perception, hang2023efficient} found that different noise levels correspond to different tasks at training time. The very earliest time-levels were found to adjust imperceptible, unimportant features, while the larger time levels formed general attributes or just interpolated to the center of the target distribution \cite{Karras2022edm}. Timesteps closer to the early-middle of the interval contributed the majority of the important features and details. Thus, we prioritize chaining from these early-middle timesteps. 

Our discretization scheme warps continuous time to contain far more timesteps at the start of the time interval. Thus, we use subdivision of discretized time for our timesteps. That is, for three-step generation, we chain at timesteps $\{t_{\frac{2N}{3}}, t_{\frac{N}{3}}\}$ where $N$ is the total number of steps. This strategy achieves the desired focus on the early-middle timesteps and behaves differently from other simple strategies such as subdividing continuous time. We qualitatively validate this comparison in Section~\ref{chaining ablation}.

\subsection{Implementation Details}
We maintain design choices such as predicting action sequences from \citet{Chi-RSS-23}, with the goal of focusing our experimental evaluation on the trained network. To this end, we also maintain the 1D Convolutional UNet architecture from Diffusion Policy for our teacher model. This architecture conditions on observations and the diffusion timestep $t$ using FiLM blocks, and diffuses through the action domain using 1D convolutional blocks. 

For our student model, we use the same architecture except with expanded FiLM blocks to accomodate conditioning on the stop timestep, $s$. We warm start our student model with the trained teacher model and we zero initialize these expanded FiLM layers to prevent them from delaying the warm started parameters effectiveness, allowing for faster convergence. 

Diffusion Policy also includes results for a Diffusion Transformer as opposed to the 1D Convolutional UNet. We choose to use the UNet due only to Diffusion Policy's \cite{Chi-RSS-23} remark that the transformer often required more hyperparameter tuning than the UNet: the choice of architecture is orthogonal to our method and Consistency Policy should benefit from a properly tuned transformer backbone just as Diffusion Policy did. 

\section{Experiments}
We begin by demonstrating Consistency Policy's strengths in accuracy and inference speed on a variety of common robotics baselines that include both image and state based control over short and long horizon tasks. We then deploy Consistency Policy in a real world trash cleanup task and compare our method against baselines in a compute constrained environment. Finally, we perform ablations over our core design choices and explore the intricacies of our model. 

\subsection{Baselines}
Since our goal is for Consistency Policy to maintain Diffusion Policy's performance while reducing inference time, the DDPM and DDiM variants of Diffusion Policy are our most important baselines. Additionally, ParaDiGMS \cite{shih2023parallel} attempts to increase the generation speed of Diffusion Policy by parallelizing the denoising process. The authors publish average speed-ups for DDPM and DDiM, the two diffusion schedulers used in Diffusion Policy, of 3.7x and 1.6x respectively. 
ParaDiGMS's experiments do not show any degradation in performance when using parallel sampling, but they do assume access to sufficient compute. Thus, we construct an optimistically strong baseline by assuming these speedups can be realized without degrading performance from the standard sequential samplers. 

As mentioned previously, we maintain the UNet architecture and Diffusion Policy infrastructure (such as the image encoder and normalization) between all experiments and methods. We also use Diffusion Policy's input and output formats across all methods. Specifically, we take as input two frames of observations (including wrist camera image and third person view camera image, and end effector pose) and output a sequence of end effector poses. Doing so allows us to directly compare the generation speed and success rates of the baselines versus our own. 

\begin{figure}
    \centering
        \includegraphics[width=\linewidth]{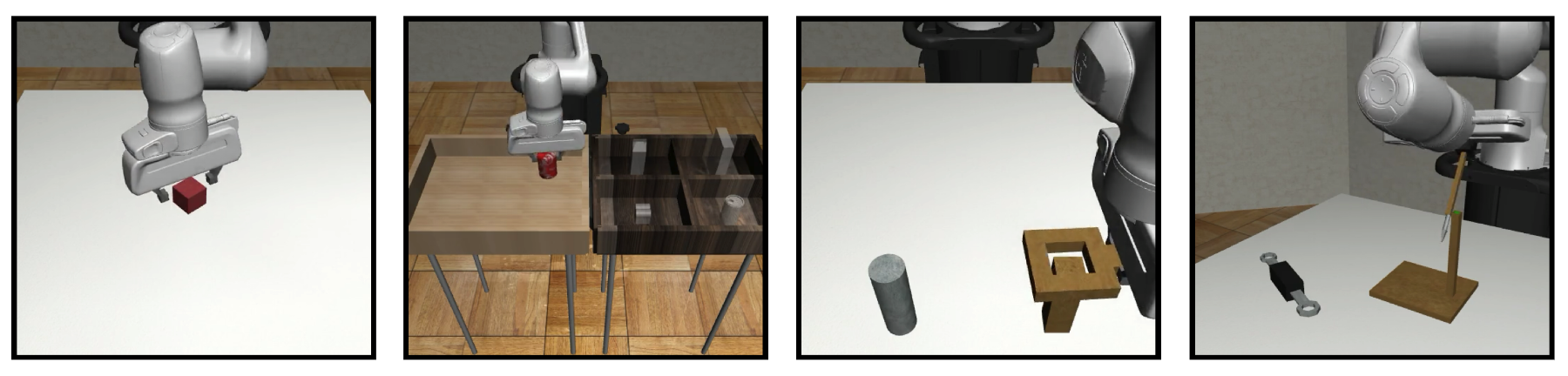}
    \caption{\textbf{Robomimic Tasks}. We evaluate our method on the single-robot Robomimic \cite{mandlekar2021matters} tasks. From left to right, and in increasing order of difficulty, we test Lift, Can, Square, and Tool Hang. }
    \label{fig:sim-exps}
\end{figure}

\subsection{Simulation Experiments}
\textbf{Tasks: }
We evaluate Consistency Policy on six tasks across three benchmarks \cite{mandlekar2021matters, florence2021implicit, gupta2020relay}. 
Robomimic, Push-T, and Kitchen are all standard benchmarks for visuomotor and state-based policy learning, and were tested in the original Diffusion Policy \cite{Chi-RSS-23} and ParaDiGMS \cite{shih2023parallel} papers. 

\begin{enumerate}
    \item \textit{Robomimic}: From the robomimic \cite{mandlekar2021matters} benchmark suite, we evaluate our method on the Lift, Can, Square and Tool Hang tasks, which compromise all the single-robot tasks in Robomimic. For each task, we report results for policies using image-based observations and the proficient human demonstration dataset, which contains 200 demonstrations per task.
    \item \textit{Push-T}: Adapted from IBC \cite{florence2021implicit}, push-T involves pushing a T-shaped block to a fixed target using a circular end-effector. We use a dataset of 200 expert demonstrations from \cite{Chi-RSS-23} and report results for policies using \textbf{state-based observations}.
    \item \textit{Franka Kitchen}: Proposed in \cite{gupta2020relay}, this state-based task involves completing four tasks in a kitchen environment in any order. We use a human demonstration dataset of 566 demonstrations and report results for policies using state-based observations. This task is specifically useful for testing \textbf{long-horizon capabilities}, as it consists of many sub-tasks that have to be completed separately.  
\end{enumerate}

\begin{figure}
    \centering
        \includegraphics[width=\linewidth]{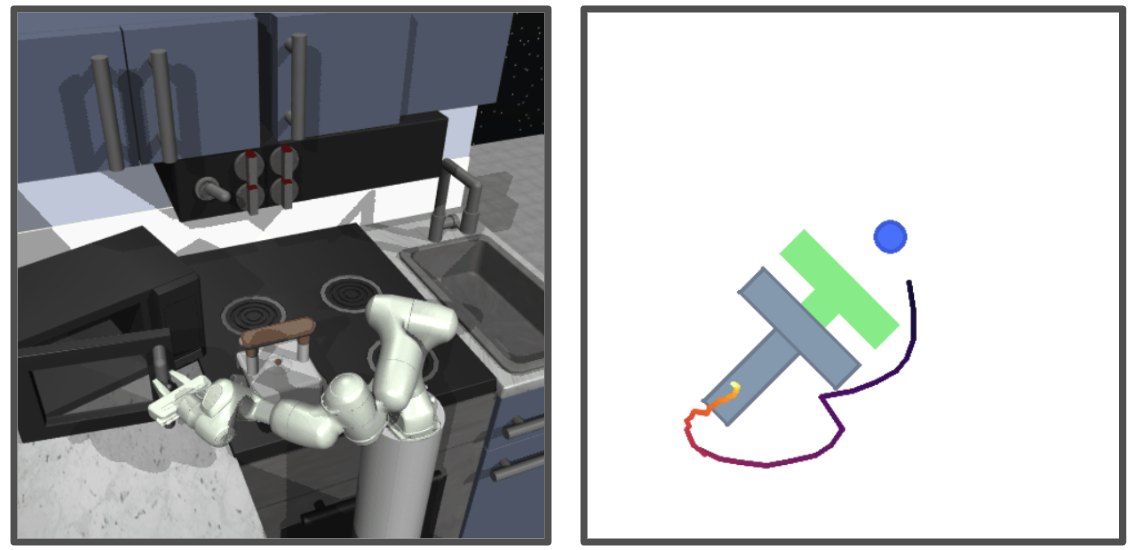}
    \caption{\textbf{State-based Simulation Tasks}. We also evaluate our method on two state-based tasks: Franka Kitchen \cite{gupta2020relay} (Left) and Push-T \cite{florence2021implicit, Chi-RSS-23} (Right). Franka Kitchen requires long-horizon and multi-stage performance over a variety of tasks that can be done in any order, while Push-T tests contact-rich manipulation of a T-block using a point force.}
    \label{fig:state-sim-exps}
\end{figure}

\textbf{Metrics:} The key metric we report in the Robomimic experiments is the \textbf{average success rate} earned by a particular policy network on the given task, along with the standard error of this metric. 
We adopt the procedure from ParaDiGMS~\cite{shih2023parallel} and compute averages and standard errors using the best checkpoint evaluated 200 times in an online setting. Push-T reports the percentage of the target area which is covered, and is otherwise measured in the same way as the previous tasks. 
\label{eval-method}
The other important metric that varies between methods is generation time, which we measure using the Number of Function Evaluations (NFE). Since inference cost for these models is dominated by NFE and the network architectures are held constant, NFE provides a good estimate of relative performance unbiased by GPU imbalances. For DDPM and DDiM, we take the speedups from ParaDiGMS \cite{shih2023parallel} into account by dividing the NFE we used by ParaDiGMS's reported speedup. Diffusion Policy \cite{Chi-RSS-23} evaluated their models with 100 steps of DDPM and 15 steps of DDiM, so we report $\frac{100}{3.7}=27$ and $\frac{15}{1.6}=9$ NFE respectively, rounding down in both cases.

\textbf{Results:}
Table ~\ref{tab:average_reward_combined} displays results for the DDPM and DDiM baselines as well as single-step and multi-step Consistency Policy, which are our contributions. For DDiM on Push-T, we use the result reported in ParaDiGMS \cite{shih2023parallel}. All other metrics were computed by us. 

\begin{table}[!t]
\centering
\resizebox{\linewidth}{!}{
\begin{tabular}{@{}llccccc@{}} 
\toprule
Policy & NFE & Lift & Can & Square & ToolHang & Push-T \\ 
\midrule
DDPM        & 27 & 1.00 & .97 $\pm$ .01 & .93 $\pm$ .02 & .79 $\pm$ .03 & .87 $\pm$ .03 \\
DDiM        & 9  & 1.00 & .82 $\pm$ .03 & .85 $\pm$ .03 & .14 $\pm$ .02 & .78 $\pm$ .03 \\
CP (ours)   & 1  & 1.00 & .98 $\pm$ .01 & .92 $\pm$ .02 & .70 $\pm$ .03  & .82 $\pm$ .03 \\
CP (ours)   & 3  & 1.00 & .95 $\pm$ .02 & .96 $\pm$ .01 & .77 $\pm$ .03 & .84 $\pm$ .03 \\
\bottomrule
\end{tabular}}
    \vspace{3mm} 
\caption{\textbf{Simulation Benchmark Results} -- Results presented are average success rates (for Robomimic Tasks) as well as target area covered (for Push-T) averaged over 200 rollouts. Each method has an associated Number of Function Evaluations (NFE) metric that dominates its runtime. Note that we are optimistic in assuming that speeding up the baseline DDPM and DDiM Policies \cite{Chi-RSS-23} with ParaDiGMS \cite{shih2023parallel} does not result in a reduction of success rates. We report NFEs for the DDPM and DDiM Policies by dividing the original NFEs of 100 by the speedups reported in ParaDiGMS. }
\label{tab:average_reward_combined}
\end{table}

Consistency Policy (CP) showcases strong single and multi step performance across all tested tasks. Single-step CP often falls in between DDPM and DDiM in terms of success rate, especially on the harder tasks such as Square and Tool Hang, but is at least an order of magnitude faster than both. 3-step CP outperforms single-step CP and is competitive with DDPM in terms of accuracy and is 3 and 9 times faster than DDiM and DDPM, respectively.

On Robomimic Can, single-step CP actually outperforms 3-step CP and registers a marginal improvement over DDPM. This divergence can be explained by stochasticity on an easy task: if the first CP generation is already earning .98 success rate, subsequent chaining steps may not have much room to improve outputs and can instead worsen performance. This explains why 3-step CP outperforms single step generation so heavily on Tool Hang: since the task is harder, the chained outputs can be substantially better than the single step outputs. 

\begin{table}[H]
\centering
\caption{\textbf{Franka Kitchen Simulation Results} -- we measure results on Franka Kitchen as in Diffusion Policy \cite{Chi-RSS-23}, with p$x$ denoting the frequency of interacting with $x$ or more objects. Franka Kitchen is a state-based task that tests both multi-stage and long-horizon performance. As in Table \ref{tab:average_reward_combined}, NFE's for DDPM and DDiM are divided by the speedups reported in ParaDiGMS \cite{shih2023parallel}.}
\label{tab:kitchen}
\begin{tabular}{@{}llcccc@{}} 
\toprule
Policy & NFE & p1 & p2 & p3 & p4\\ 
\midrule
DDPM      & 27 & 1.00 & 1.00 & 1.00 & .98 $\pm$ .01 \\
DDiM    & 9 & 1.00 & .98 $\pm$ .01 & .98 $\pm$ .01 & .93 $\pm$ .02\\
CP (ours)  & 1 & .99 $\pm .01$& .96 $\pm .01$ & .95 $\pm .02$& .93 $\pm .02$\\
CP (ours)  & 3 & .99 $\pm .01$& .96 $\pm .01$ & .97 $\pm .01$& .94 $\pm .02$\\
\bottomrule
\end{tabular}
\end{table}

Results for the Franka Kitchen task \cite{gupta2020relay} are presented in Table \ref{tab:kitchen}. Single-step Consistency Policy returns strong results for the first two stages of this task but struggles more in the later stages. More exploration in long-horizon environments is required to understand what exactly Consistency Policy struggles to learn in this longer task. 

Table \ref{tab:sim_inference_speed} showcases wall clock times for each of the policies in simulation, specifically over the Robomimic Square task. As expected, Consistency Policy completes inference orders of magnitude faster than the Diffusion Policy baselines.  
\begin{table}[h]
\centering
\begin{tabular}{@{}lcc@{}} 
\toprule
Policy & NFE & Inference Time (ms) \\ 
\midrule
DDPM        & 100 & 110 \\
DDiM        & 15 & 11 \\ 
CP (ours)   & 1 & 1 \\
CP (ours)   & 3 & 2 \\
\bottomrule
\end{tabular}
    \vspace{3mm} 
\caption{\textbf{Simulation Inference Speeds} -- Simulation inference speeds were measured on an NVIDIA P5000 datacenter GPU and averaged over 50 rollouts. Benchmarking was done with vanilla Diffusion Policy since we used this as our baseline.}
\label{tab:sim_inference_speed}
\end{table}
\subsection{Real World Experiments}
\label{real-world-exp}
\textbf{Tasks}: We evaluate Consistency Policy in the real world on three tasks: Trash Clean Up, Plug Insertion, and Microwave.

\begin{figure}
    \centering
        \includegraphics[width=\linewidth]{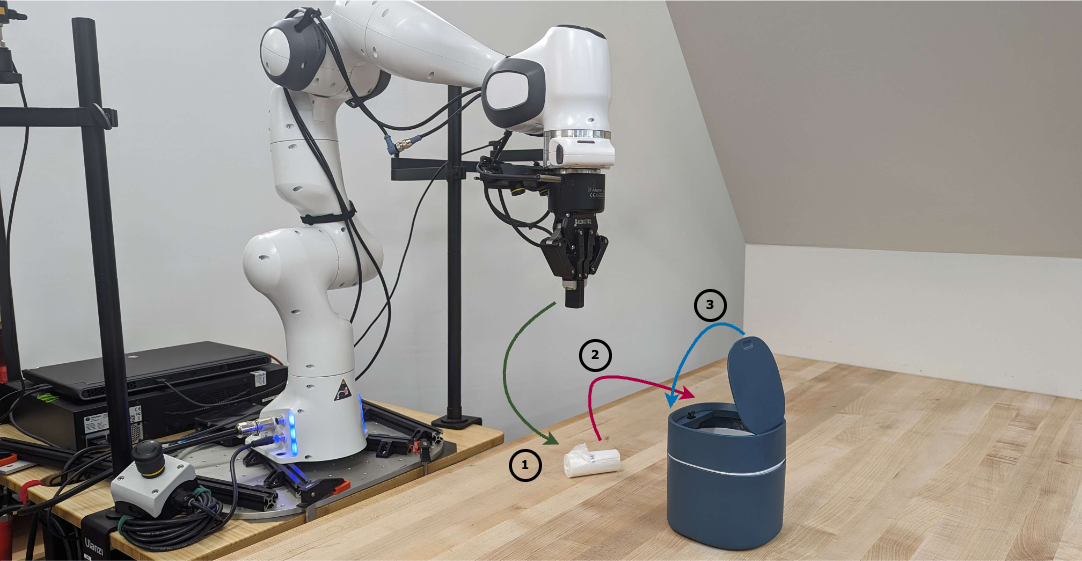}
    \caption{\textbf{Rubbish Clean Up}. This task involves: (1) picking up trash, (2) placing the trash in the trash can, then (3) closing the lid of the trash can.}
    \label{fig:real-world}
\end{figure}

\begin{figure}
    \centering
        \includegraphics[width=\linewidth]{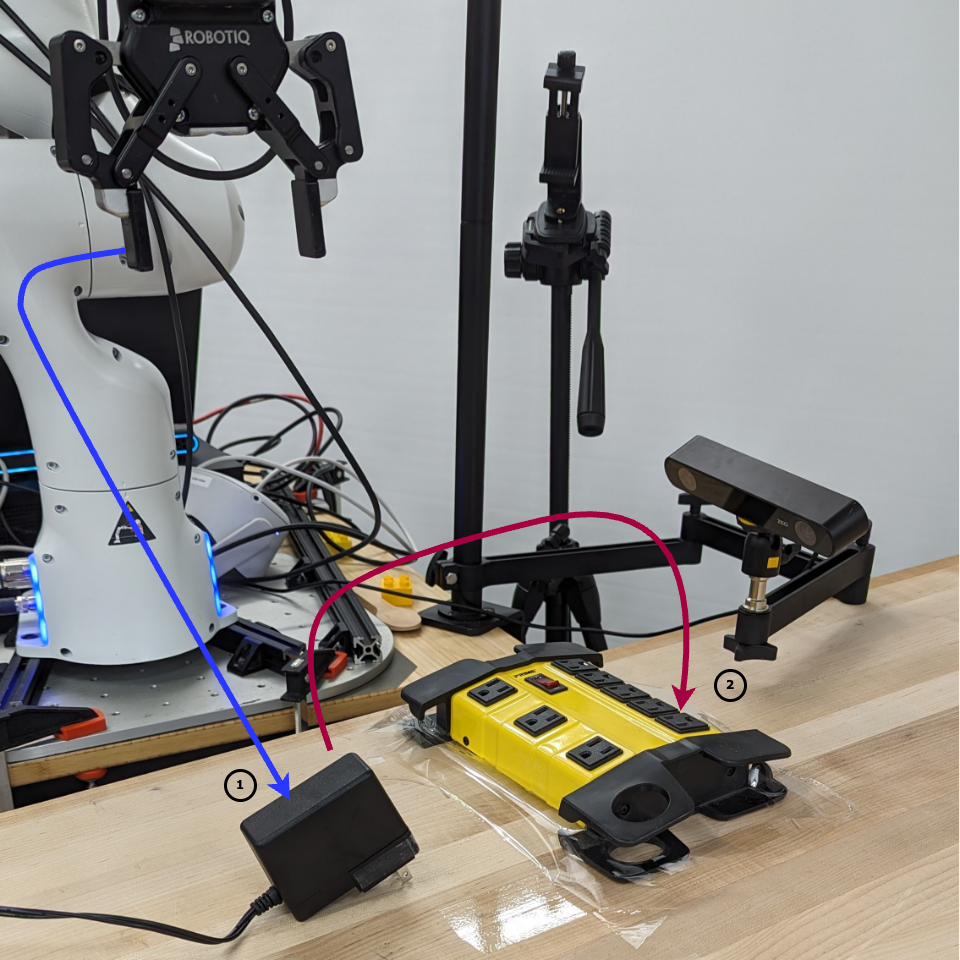}
    \caption{\textbf{Plug Insertion}. This task involves: (1) picking up a power adapter, (2) bringing it to a plug and inserting it completely}
    \label{fig:real-world-plug}
\end{figure}

\begin{figure}
    \centering
        \includegraphics[width=\linewidth]{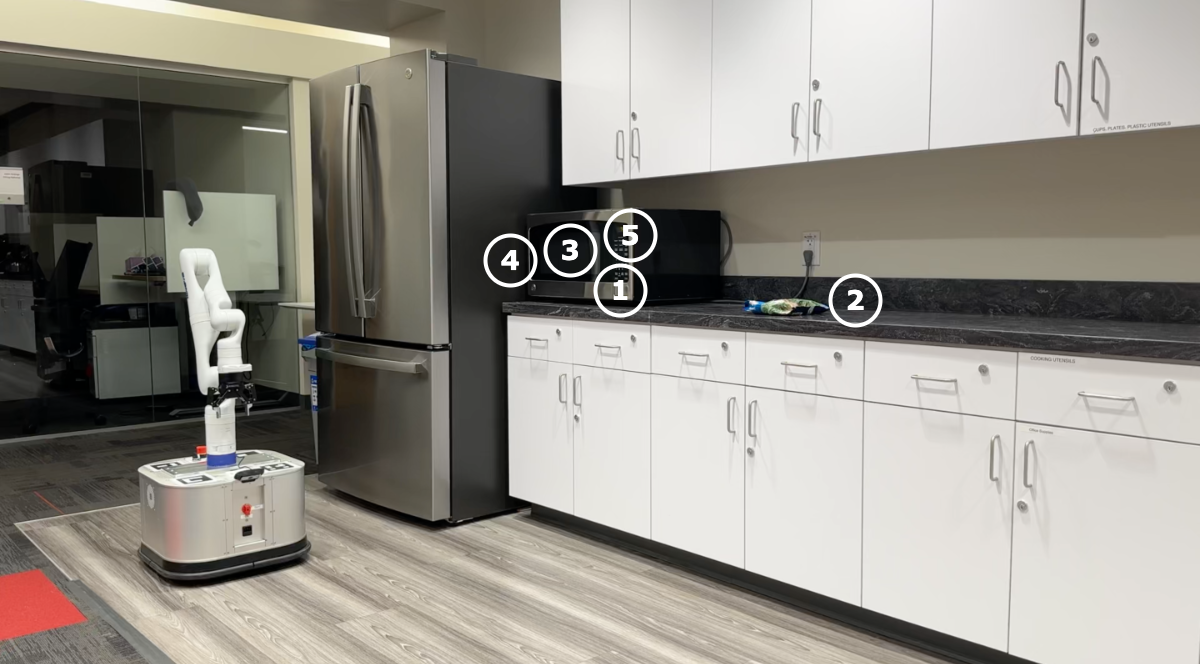}
    \caption{\textbf{Microwave}. This task involves: (1) navigating to and opening a microwave, (2) navigating to and picking up a bag of broccoli, (3) placing the bag inside the microwave, (4) closing the microwave door, and (5) starting the microwave.}
    \label{fig:real-world-microwave}
\end{figure}

\begin{enumerate}
    \item \textit{Trash Clean Up}: The robot has to pick up trash lying near the can, place the trash inside of the can, and close the can's lid (see Fig. \ref{fig:real-world}). 
    \item \textit{Plug Insertion}: The robot has to pick up a plug and insert it into a socket. This task tends to be more contact rich and requires precision (see Fig. \ref{fig:real-world-plug})
        \item \textit{Microwave}: The robot (a mobile manipulator) has to open a microwave, navigate to and retrieve a bag of broccoli, place it in the microwave, close the microwave door, and press the ``vegetable'' button to start the microwave. This task tests long horizon performance and control of a mobile base along with a standard robot arm (see Fig. \ref{fig:real-world-microwave}).
\end{enumerate}
For the first two tasks, we use a laptop containing a single 3070 Ti GPU with 8GB of VRAM for inference. Note that we are unable to run ParaDiGMS effectively on this setup due to ParaDiGMS's high VRAM usage. We also note that, on the 3070 Ti GPU, the 100-step DDPM variant of Diffusion Policy has an inference time of around 1.5 seconds per forward pass. Thus, we choose as our baseline method the faster and more realistic DDiM variant of Diffusion Policy, which uses 15 steps for policy inference. 

We maintain the same policy inputs, outputs, and architecture as in the simulation experiment setup, except the observation size of the plug insertion task: for this task, we use image size of 256x256. At a given timestep $t$, our policy receives: an over-the-shoulder image, a camera-in-hand image, current end-effector pose and current gripper width. Our policy outputs an action sequence of length 16, where each step is a 10D vector of goal end effector position (3D), goal effector rotation (6-D) and goal gripper position (1D). 

For the mobile manipulation microwave task, we use a Kinova Gen3 7-DoF arm mounted on a holonomic mobile base. Inference was completed on the same laptop with a 3070 Ti GPU and transmitted to the robot over a router connection. 84x84 resolution image observations were taken from a wrist camera as well as a base mounted camera. Low dimensional observations include 3D base pose, 3D arm position, 4D arm rotation quaternion, and 1D for gripper width. Individual actions have dimension 13 (3 for the base pose, 3 for the arm position, 6D arm rotation, 1D gripper) and are chunked together temporally.

\textbf{Metrics:} For each trained policy, we report \textbf{average success rates} on one policy checkpoint (selected using validation mean-squared error). We also report inference time on the 3070 Ti GPU. For success rates, we average over 10 trials for the first and third tasks while we average over 20 trials for the second task. Starting positions were randomized between trials for the first two tasks and were static for the third task. 

\textbf{Results:} Table \ref{tab:real_world_results} shows how the baseline DDiM-variant of Diffusion Policy achieves similar average success rates as our method on the Rubbish Clean Up and Plug Insertion task. However, our method has much lower inference time ($\sim$9x lower latency). Our method maintains its inference speed advantage in the Microwave task and performs slightly worse than DDiM. More discussion about the mobile task in particular is present in Limitations {see Sec. \ref{limitations}}.

As Consistency Policy requires 15x fewer forward passes than DDiM, one might expect the full inference speed-up to also be around 15x instead of around 9x. This discrepancy exists because of overhead costs such as the observation encoder. We discuss inference speeds further in the Appendix.

\begin{table}[!t]
\centering
\begin{tabular}{@{}l@{\hskip 3pt}c@{\hskip 3pt}c@{\hskip 3pt}c@{\hskip 3pt}c@{\hskip 3pt}c@{}}
\toprule
 & \multicolumn{2}{c}{Trash Clean Up} & \multicolumn{2}{c}{Plug Insertion} & \multicolumn{1}{c}{Microwave} \\
\cmidrule(lr){2-3} \cmidrule(lr){4-5} \cmidrule(lr){6-6}
       & \multicolumn{1}{l}{Success} & \multicolumn{1}{l}{Inference}  & \multicolumn{1}{l}{Success} & \multicolumn{1}{l}{Inference} & \multicolumn{1}{l}{Success Rate}\\
       & \multicolumn{1}{l}{Rate}   & \multicolumn{1}{l}{Time (ms)} & \multicolumn{1}{l}{Rate}   & \multicolumn{1}{l}{Time (ms)} & \multicolumn{1}{l}{}\\
\midrule
DDiM   & 0.8 $\pm$ .13 & 192  &  0.6 $\pm$ 0.11 & 198 & 0.5 $\pm$ 0.16 \\
CP (ours) & 0.8 $\pm$ .13 & 21 & 0.7 $\pm 0.10$ & 22 & 0.4 $\pm$ 0.15\\
\bottomrule
\end{tabular}

\vspace{3mm}
\caption{\textbf{Real World Experiment Results for Multiple Tasks} -- Results presented for Consistency Policy and the DDiM variant of Diffusion Policy. Success rates and standard errors are presented for each task. Inference speeds are also presented for the first two tasks.}
\label{tab:real_world_results}
\end{table}

\subsection{Ablations}

We perform several ablations to validate and explore our design choices. Unless otherwise stated, we calculate success rates using the same evaluation methodology used in our simulation experiments (see Sec.~\ref{eval-method}). We choose Robomimic Square and Toolhang for these experiments since these were the two hardest image-based tasks. 

\textbf{Consistency Objective}: Recall that the CTM framework learns the function $g_\theta(\mathbf{x}_t, t, s)$ by in part optimizing the consistency objective in Eq \ref{CTM Loss}. There are three natural consistency objectives one could optimize, which differ in the choices of starting points $t, u$ as well as the choice of stop point $s$ - all visualized in Fig~\ref{fig:training}. \citet{song2023consistency} proposed the Consistency Distillation objective that enforces consistency between adjacent points $t$ and $u$ with $s=0$. \citet{kim2023consistency} proposed the non-adjacent CTM objective that enforces consistency between any points $t$ and $u$ denoised down to any $s < u < t$. Finally, one can use adjacent $t$ and $u$ like Consistency Distillation \cite{song2023consistency} (also termed `local consistency' from CTM \cite{kim2023consistency}) as well as arbitrary $s$. In our experiments, we found that this third objective worked the best.
In our ablation (see Table \ref{tab:consistency-objective-ablation}) comparing all three objectives, we vary the consistency objective but maintain the auxillary DSM objective as in Eq. \ref{total loss}. 


\begin{table}[H]
\centering
\caption{Consistency Objective Ablation On Square Task}
\label{tab:consistency-objective-ablation}
\begin{tabular}{@{}lc@{}} 
\toprule
Method & Success Rate\\ 
\midrule
Consistency Distillation & .88 $\pm$ .02 \\
CTM      & .91 $\pm$ .02  \\
CTM-local (ours)  & .92 $\pm$ .02 \\
\bottomrule
\end{tabular}
\end{table}

CTM and CTM-local have similar success rates, though both outperform Consistency Distillation by a slight margin. However, CTM is far more computationally expensive to train than CTM-local and Consistency Distillation because of the multiple teacher denoising steps that are required to move from $t \rightarrow u$. Even with the constraint that $t-u \leq 10$, 
we found that CTM trained more than $40\%$ slower than Consistency Distillation and CTM-local, which both train at the same speed. These results were measured on an NVIDIA RTX A5000.

\textbf{Initial Sample Variance}: Diffusion requires sampling an initial position from a Gaussian which is then denoised into a valid action prediction. The normalized initial position is traditionally sampled from the unit Gaussian, $\mathcal{N}(0, 1)$. We chose to instead sample from a low-variance Gaussian, $\mathcal{N}(0, \frac{1}{T^2})$, to push the output to remain more in-distribution. 
\label{initial-var-ablation}

In Table~\ref{tab: initial variance}, we compared the results from sampling from the high variance (original) versus low variance (ours) initial position using both the single-step and 3-step Consistency Policy method on the Robomimic Square task.

\begin{table}[H]
\centering
\caption{Initial Sample Variance Ablation On Square Task}
\label{tab: initial variance}
\begin{tabular}{@{}lcc@{}} 
\toprule
Initial Variance & 1-step & 3-step \\ 
\midrule
1        & .9 $\pm$ .02 & .91 $\pm$ .02 \\
$\frac{1}{T^2}$       & .92 $\pm$ .02  & .96 $\pm$ .01  \\

\bottomrule
\end{tabular}
\end{table}

Though the low variance inital sample outperforms with both methods, it seems to serve a much larger impact in the multi-step case. This is potentially because of the noising that occurs between steps. The benefit of higher initial variances is expressivity and multimodality over a complex data distribution: while the first step begins at some $\mathbf{x}_t \sim \mathcal{N}(0, \frac{1}{T^2})$, and so will end at some more central end point, the subsequent noising and denoising steps might preserve the high-variance position's expressivity.



\textbf{Preset Chaining Steps:} \label{chaining ablation}
To validate our decision of focusing chaining to early-middle timesteps by subdividing discretized time rather than continuous time, we tested both of these chaining procedures on Square and Tool Hang. Both experiments are reported in Table~\ref{tab:chaining steps} and were done with 3 chaining steps and even subdivisions of the underlying time space (the discretized mesh versus the continuous time interval). 

\begin{table}[H]
\centering
\caption{Chaining Steps Ablation on Square and Tool Hang}
\label{tab:chaining steps}
\begin{tabular}{@{}lcc@{}} 
\toprule
NFE & Square & Tool Hang \\ 
\midrule
Discretized      & .96 $\pm$ .01 & .77 $\pm$ .03 \\
Continuous     & .94 $\pm$ .02 & .72 $\pm$ .03  \\
\bottomrule
\end{tabular}
\end{table}

Discretized subdivisions outperformed continuous subdivisions heavily on Tool Hang, while both methods achieved similar results on Square (with discretized subdivisions doing slightly but not significantly better). As in other ablations, we think the benefit of this choice is most apparent on harder tasks such as Tool Hang where there is more room for improvement. We suggest that any user wishing to improve performance on a difficult task begin by trying subdivided discretized time and only attempt further hyperparameter tuning if they still need to do so. 

\textbf{Teacher Model Quality:}
Since Consistency Policy requires distilling a pretrained teacher model into a student network, it is relevant to understand how important the teacher model's performance is to the eventual performance attained by the student model. We tested distillation using three different teacher models of varying quality against the Square task and report results in Table~\ref{tab:teacher quality}. 

\begin{table}[H]
\centering
\caption{Robustness to Teacher Model Quality on Square Task}
\label{tab:teacher quality}
\begin{tabular}{@{}lcc@{}} 
\toprule
Teacher Success Rate & Student Success Rate\\ 
\midrule
.92 $\pm$ .02     & .92 $\pm$ .02 \\
.88 $\pm$ .03     & .92 $\pm$ .02   \\
.84 $\pm$ .03    & .88 $\pm$ .03 \\
\bottomrule
\end{tabular}
\end{table}

While there was a slight correlation observed between teacher quality and student success rate, Consistency Policy maintains robustness against the teacher's success rate over this range of teacher qualities. While the consistency objective $\mathcal{L}_{CTM}$ (Eq. \ref{CTM Loss}) depends directly on the teacher, the DSM objective $\mathcal{L}_{DSM}$ (Eq. \ref{DSM loss}) is independent of the teacher and is likely able to maintain student performance even as the teacher gets worse. This bodes well for deployment in real world tasks where extensive testing of the teacher model might not be possible.

\textbf{The role of dropout in the CTM Objective:}  \label{regular ablation}
Regularization techniques such as dropout \cite{srivastava2014dropout} are usually used to prevent a highly expressive model from overfitting on the training dataset. However, in our experiments, we found that dropout plays a far more important role than expected in the CTM objective. 

Recall that after both $t \rightarrow s$ and $u \rightarrow s$ are computed, the resulting predictions are brought back to time 0 before the loss is calculated (see Eq. \ref{CTM Loss} and Fig.~\ref{fig:training}). 
It is our hypothesis that when $g_\theta$ reaches a certain level of performance, such as when it is warm started with parameters from a pretrained teacher model, the loss $\mathcal{L}_{CTM} = d(g_\theta(\mathbf{x}_s^{(t)}, s, 0; o), g_\theta(\mathbf{x}_s^{(u)}, s, 0; o))$ vanishes, providing no training signal. 

The DSM loss (Eq. \ref{DSM loss}) that EDM uses to train can be interpreted as teaching the diffusion model to take an original sample $\mathbf{x}$ noised using \textbf{any} noise sample $\epsilon$ and some timestep $s$ (we choose this notation deliberately) back to the original sample in a single step. Intuitively, feeding a strong image-space diffusion model two noised versions of the same image should return similar images in both cases, even if the two noised version do not lie on the same trajectory (meaning they were not formed with the same sampled $\epsilon$). Following this reasoning, if $\mathbf{x}_s^{(t)}$ and $\mathbf{x}_s^{(u)}$ are good approximations of $\mathbf{x}$ noised to time $s$ using any $\epsilon$, we should expect the outputs $g_\theta(\mathbf{x}_s^{(t)}, s, 0; o)$ and $g_\theta(\mathbf{x}_s^{(u)}, s, 0; o)$ to be very similar even if $d(\mathbf{x}_s^{(t)}, x_s^{(u)})$ is large. And indeed, empirically we found that $d(\mathbf{x}_s^{(t)}, \mathbf{x}_s^{(u)})$ was at least two orders of magnitude larger than $d(g_\theta(\mathbf{x}_s^{(t)}, s, 0; o), g_\theta(\mathbf{x}_s^{(u)}, s, 0; o))$ with dropout disabled. 

When dropout is enabled, the $s \rightarrow 0$ step stops being deterministic in this manner. Because the Consistency Network can no longer rely on these later steps to bring points from different trajectories closer together, there is more signal acting to directly enforce self-consistency on  $d(\mathbf{x}_s^{(t)}, \mathbf{x}_s^{(u)})$ as opposed to making $\mathbf{x}_s^{(t)}$ and $\mathbf{x}_s^{(u)}$ good predictors of any $\textbf{x}_s$. 

As an initial step towards exploring this hypothesis, we removed dropout from only the two generations from $s \rightarrow 0$ at training time while retaining it throughout the rest of the network. 
\begin{table}[H]
\centering
\caption{Effect of Removing $s \rightarrow 0$ Dropout on Square Task}
\label{tab:dropout-ablation}
\begin{tabular}{@{}lc@{}} 
\toprule
Dropout & Success Rate \\ 
\midrule
Enabled        &  .92 $\pm$ .02\\
Disabled      & .86 $\pm$ .03  \\

\bottomrule
\end{tabular}
\end{table}
As seen in Table \ref{tab:dropout-ablation}, removing dropout in just this part of training resulted in decreased success rate, which is to be expected if dropout is indeed responsible for most of the signal from the consistency objective. For this, as well as our other results, dropout was set to 0.2. 
\\ \textbf{Consistency Training:}
\cite{anonymous2024consistency} is a concurrent work in RL that learned a state-based policy using Consistency Training \cite{song2023consistency, song2023improved}, which substitutes the trained teacher model used to calculate $\mathbf{x}_u$ from $\mathbf{x}_t$ (for $u=t-1$) with a Monte Carlo estimate of the score function. We implemented this system (which we refer to in Table \ref{tab:consistency training} as CT Policy) and tested it on Robomimic Lift and Square with single-step generation. Our method's results on Lift and Square are displayed as well for comparison. 

\begin{table}[H]
\centering
\caption{Consistency Training Ablation on Lift and Square}
\label{tab:consistency training}
\begin{tabular}{@{}lcc@{}} 
\toprule
NFE & Lift & Square \\ 
\midrule
CT Policy     & .91 $\pm$ .02 & .55 $\pm$ .04 \\
CP (ours)     & 1.0 &  .92 $\pm$ .02 \\
\bottomrule
\end{tabular}
\end{table}

\section{Limitations}
\label{limitations}
While Consistency Policy marks strong improvements in raw speed over Diffusion Policy while retaining performance, the distillation procedure does carry drawbacks. Some of Diffusion Policy's \cite{Chi-RSS-23} benefits include its ability to represent multimodality in the action distribution, and stable training. Consistency Policy makes trade-offs regarding these two attributes. 

Diffusion Policy's multimodality likely comes from DDPM's integration over a \emph{Stochastic} Differential Equation, as opposed to the Ordinary Differential Equation that EDM and CTM learn to integrate -- the consistency objective requires distillation of a deterministic trajectory. Indeed, we found that both the teacher EDM policy and Consistency Policy lose some of Diffusion Policy's multimodality, e.g., by favoring one side of the Push-T task over the other. However, Consistency Policy still performs well on the Push-T task, suggesting that this lack of multi-modality is not hurting us on the standard evaluation tasks used by related work. In future work, we will explore how we can potentially re-introduce multimodality to Consistency Policy through more complex sampling schemes. 

We also noticed that Consistency Policy is slightly less stable during training than Diffusion Policy, likely due to the self-referential nature of the consistency objective (see Eq. \ref{CTM Loss}). 

Our mobile manipulation results, where Consistency Policy slightly underperforms Diffusion Policy on accuracy, showcases the varied strengths and weaknesses of our method.

Note first that we were unable to train Consistency Policy on the Microwave task until it was no longer improving due to external time constraints. Note as well that Consistency Policy empirically needs more time in training to reach the same level of performance as Diffusion Policy, especially when taking into account time taken to train the teacher policy. This usually means both more epochs as well as more time per epoch since a Consistency Policy training step requires running the teacher model in addition to multiple forward passes of the student network.

This increase in training time is exacerbated by the difficulty of the task (Microwave took much longer to distill than either of the other tasks did) and is intuitive given that Consistency Policy is fitting a harder problem (a single-step policy rather than a multi-step policy over the same task). It is important to consider how these trade-offs interact with your unique use case, rather than comparing these tools in a vacuum. 
\section{Conclusion} 
\label{sec:conclusion}

We present the use of consistency-based training objectives for training high performing, low-latency visuomotor robot policies. In our evaluation suite of 9 tasks in simulation and in the real world, we demonstrate that consistency policies yield a dramatic increase in inference speeds compared to previous diffusion policy methods without sacrificing success rates. Through our ablations, we also highlight key design decisions that led to visuomotor policies with the highest success rates. These include: the choice of consistency objective, lowering the initial sample variance, the use of dropout, and the choice of preset chaining steps. In future work, we hope to explore deeper explanations behind certain experimental results, such as chaining dynamics and the role of dropout along the $s \rightarrow 0$ generations, and also port Consistency Policy to robots that can better utilize fast inference speeds, including legged, winged, and other mobile platforms. 

\textbf{Acknowledgements: }Toyota Research Institute provided funds to support this work.


\bibliographystyle{plainnat}
\bibliography{references}

\clearpage
\onecolumn



\newpage
\setlength{\parskip}{1em}

\section*{Overview}
The appendix offers additional details with respect to the inference speed benefits of Consistency Policy (Appx. \ref{inf_appendix}), the low variance initial sampler (Appx. \ref{var_appendix}), and the real world experiment (Appx.~\ref{setup_appendix}). 


\begin{appendices}


\section{Inference Speeds}
\begin{table}[H]
\centering
\begin{tabular}{@{}lccc@{}} 
\toprule
Policy & Image Encoder Time (ms) & Network Time (ms) & Total Inference Time (ms) \\ 
\midrule
DDiM        & 6 & 179 & 192 \\ 
CP (ours)   & 6 & 13.5 & 21 \\
\bottomrule
\end{tabular}
    \vspace{3mm} 
\caption{All time measurements are given in milliseconds. \textbf{Image Encoder Time} is the time spent encoding image observations before they are passed to the Policy Network. \textbf{Network Time} is the amount of time spent on forward passes of the Policy Network. \textbf{Inference Time} is the total time in between image observations being inputted and actions being deployed. This includes the previous two measurements as well as any additional time costs, such as time for shuttling data.}
\label{tab:inference_appendix}
\end{table}
In our real world experiments (Section \ref{real-world-exp}), we state that 15-step DDiM had an inference time of 192 ms while Consistency Policy had an inference time of 21 ms (see Table \ref{tab:inference_appendix}). We use inference time to refer to the entire process from when the Policy Network (either Diffusion Policy or Consistency Policy) is given image observations to when predicted actions are deployed to the robot's controller. While the majority of this time is spent on forward passes of the Policy Network, some of it is also spent encoding the image observations and shuttling memory. Thus, we also measured the time spent only on forward passes of the Policy Network and the time spent on encoding the image observations. These values are reported in Table \ref{tab:inference_appendix} and were all measured on an NVIDIA 3070 Ti, which is an underpowered laptop GPU. 
\label{inf_appendix}

While the relative speed-up of inference time between DDiM and Consistency Policy is 9x, the relative speed-up of network time is 13.3x, closer to the 15x reduction in steps between our method and DDiM. 

We reported the overall inference speed increase from Section \ref{real-world-exp} for easy comparison with other speed-up methods. However, the actual inference speed-up realized by a practitioner will depend on the other systems they run alongside their Policy Network, such as their own observation encoders or parallel processes. As the Policy Network consumes a larger and larger portion of computation relative to other processes at inference time, our method will grow more impactful. 

\section{Low Variance Initial Samples}
\label{var_appendix}

One of our ablations (see Section \ref{initial-var-ablation}) found that the lower variance initial sampler improved generation quality over the higher variance initial sampler standard to EDM and Consistency Models. 

\begin{figure}[h]
    \centering
    \begin{minipage}{0.45\linewidth}
        \includegraphics[width=\linewidth]{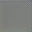}
        \caption{Low initial variance generation. The diffusion model evidently did not learn to predict the score in this region.}
        \label{fig:low_var}
    \end{minipage}
    \hfill
    \begin{minipage}{0.45\linewidth}
        \includegraphics[width=\linewidth]{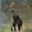}
        \caption{High initial variance generation. The same diffusion model has learned support for this higher variance region.}
        \label{fig:high_var}
    \end{minipage}
\end{figure}

Testing EDM on unconditional image generation suggests that this low-variance sample might have some specific benefit for the robotics domain. Figures \ref{fig:low_var} and \ref{fig:high_var} are generations from the same EDM model (trained on CIFAR-10, a $32$x$32$ px image dataset) using low initial variance sampling and high initial variance sampling. 

These are just heuristic results since the EDM was not trained until convergence, but it is clear that the low-variance region has no learned support as compared to the high variance initial position, which causes the low variance generation to produce the gray block in Figure \ref{fig:low_var}. 

It is possible that the difference in support for low variance regions between these domains comes from the difference in dimensionality of the data distributions the respective diffusion models are trying to fit. The manifold hypothesis states that real world datasets, such as robot actions or images, are contained in low-dimensional manifolds of $R^n$. It is intuitive, though not certain, that the manifold robot actions lie on for a given task and robot is much lower in dimensionality than the space of all CIFAR-10 images, which contain 60000 images over 10 classes and lives in the 32x32x3 pixel space. For comparison, our robot action data had dimensionality $h$ x $10$, where $h$ was the action horizon length and each action was represented by a 9D pose vector and 1D gripper action. It is possible that the score model learns to support low-variance initial positions only on sufficiently low-dimensional data manifolds, like that of robot actions, while this region is left without support when the score model is trained on a higher dimensional space. Such behavior might arise from the fact that higher dimensional Gaussians concentrate more of their mass at their edges rather than at their centers. If the Diffusion/Consistency Policy is learning a map to or from a Gaussian in N dimensions, it may support the center of that Gaussian less and less as N increases.

\section{Real World Experiment Setup}
\label{setup_appendix}
We perform the stationary arm real world experiments (Trash Pick Up and Plug Insertion) using a Franka Panda robot. We mount a wrist (Zed Mini) camera and an over-the-should camera (Zed Mini) to obtain two images at each timestep. We use an Meta Quest 2 VR setup to collect 180 demonstrations for our task. The Franka robot accepts control commands at 1kHz, and receives new commands from either the VR controller or trained policy at 15 Hz. Note that Consistency Policy generates an end effector action sequence output in around 21ms, and we supply each waypoint to the robot at 15Hz. When the policy performs inference, we send a control command to the robot to maintain its current end effector pose.

\end{appendices}

\end{document}